%% file: main.tex
\PassOptionsToPackage{table}{xcolor}
\documentclass[nohyperref]{article}

\usepackage{microtype}
\usepackage{graphicx}
\usepackage{subfigure}
\usepackage{booktabs} 

\usepackage{hyperref}

\usepackage[accepted]{icml2022}

\usepackage{amsmath}
\usepackage{amssymb}
\usepackage{mathtools}
\usepackage{amsthm}
\definecolor{LightCyan}{rgb}{0.88,1,1}

\usepackage[capitalize,noabbrev]{cleveref}

\usepackage{multirow}
\usepackage[table]{xcolor}

\theoremstyle{plain}

\theoremstyle{definition}

\theoremstyle{remark}

\usepackage[textsize=tiny]{todonotes}

\icmltitlerunning{Revisiting Architecture-aware Knowledge Distillation}

\begin{document}

\twocolumn[
\icmltitle{Revisiting Architecture-aware Knowledge Distillation: \\ Smaller Models and Faster Search}



\icmlsetsymbol{equal}{*}

\begin{icmlauthorlist}
\icmlauthor{Taehyeon Kim}{equal,yyy}
\icmlauthor{Heesoo Myeong}{comp}
\icmlauthor{Se-Young Yun}{yyy}
\end{icmlauthorlist}

\icmlaffiliation{yyy}{KAIST AI, Seoul, South Korea}
\icmlaffiliation{comp}{Qualcomm Korea YH, Seoul, South Korea}

\icmlcorrespondingauthor{Taehyeon Kim}{potter32@kaist.ac.kr}
\icmlcorrespondingauthor{Se-Young Yun}{yunseyoung@kaist.ac.kr}

\icmlkeywords{Machine Learning, ICML}

\vskip 0.3in
]
\input{math_commands}



\printAffiliationsAndNotice{\icmlEqualContribution} 

\begin{abstract}
    Knowledge Distillation\,(KD) has recently emerged as a popular method for compressing neural networks. In recent studies, generalized distillation methods that find parameters and architectures of student models at the same time have been proposed. Still, this search method requires a lot of computation to search for architectures and has the disadvantage of considering only convolutional blocks in their search space. This paper introduces a new algorithm, coined as Trust Region Aware architecture search to Distill knowledge Effectively\,(TRADE), that rapidly finds effective student architectures from several state-of-the-art architectures using trust region Bayesian optimization approach. Experimental results show our proposed TRADE algorithm consistently outperforms both the conventional NAS approach and pre-defined architecture under KD training. 
\end{abstract}

\input{section/1_intro}
\input{section/2_preliminary}
\input{section/3_method}
\input{section/4_result}

\input{section/5_conclusion}

\bibliographystyle{icml2022}
\bibliography{egbib}

\appendix
\onecolumn
\input{appendix/app1}

\end{document}

%% file: math_commands.tex


\newcommand{\figleft}{{\em (Left)}}
\newcommand{\figcenter}{{\em (Center)}}
\newcommand{\figright}{{\em (Right)}}
\newcommand{\figtop}{{\em (Top)}}
\newcommand{\figbottom}{{\em (Bottom)}}
\newcommand{\captiona}{{\em (a)}}
\newcommand{\captionb}{{\em (b)}}
\newcommand{\captionc}{{\em (c)}}
\newcommand{\captiond}{{\em (d)}}

\newcommand{\newterm}[1]{{\bf #1}}

\def\figref#1{figure~\ref{#1}}
\def\Figref#1{Figure~\ref{#1}}
\def\twofigref#1#2{figures \ref{#1} and \ref{#2}}
\def\quadfigref#1#2#3#4{figures \ref{#1}, \ref{#2}, \ref{#3} and \ref{#4}}
\def\secref#1{section~\ref{#1}}
\def\Secref#1{Section~\ref{#1}}
\def\twosecrefs#1#2{sections \ref{#1} and \ref{#2}}
\def\secrefs#1#2#3{sections \ref{#1}, \ref{#2} and \ref{#3}}
\def\eqref#1{equation~\ref{#1}}
\def\Eqref#1{Equation~\ref{#1}}
\def\plaineqref#1{\ref{#1}}
\def\chapref#1{chapter~\ref{#1}}
\def\Chapref#1{Chapter~\ref{#1}}
\def\rangechapref#1#2{chapters\ref{#1}--\ref{#2}}
\def\algref#1{algorithm~\ref{#1}}
\def\Algref#1{Algorithm~\ref{#1}}
\def\twoalgref#1#2{algorithms \ref{#1} and \ref{#2}}
\def\Twoalgref#1#2{Algorithms \ref{#1} and \ref{#2}}
\def\partref#1{part~\ref{#1}}
\def\Partref#1{Part~\ref{#1}}
\def\twopartref#1#2{parts \ref{#1} and \ref{#2}}

\def\ceil#1{\lceil #1 \rceil}
\def\floor#1{\lfloor #1 \rfloor}
\def\1{\bm{1}}
\newcommand{\train}{\mathcal{D}}
\newcommand{\valid}{\mathcal{D_{\mathrm{valid}}}}
\newcommand{\test}{\mathcal{D_{\mathrm{test}}}}

\def\eps{{\epsilon}}

\def\reta{{\textnormal{$\eta$}}}
\def\ra{{\textnormal{a}}}
\def\rb{{\textnormal{b}}}
\def\rc{{\textnormal{c}}}
\def\rd{{\textnormal{d}}}
\def\re{{\textnormal{e}}}
\def\rf{{\textnormal{f}}}
\def\rg{{\textnormal{g}}}
\def\rh{{\textnormal{h}}}
\def\ri{{\textnormal{i}}}
\def\rj{{\textnormal{j}}}
\def\rk{{\textnormal{k}}}
\def\rl{{\textnormal{l}}}
\def\rn{{\textnormal{n}}}
\def\ro{{\textnormal{o}}}
\def\rp{{\textnormal{p}}}
\def\rq{{\textnormal{q}}}
\def\rr{{\textnormal{r}}}
\def\rs{{\textnormal{s}}}
\def\rt{{\textnormal{t}}}
\def\ru{{\textnormal{u}}}
\def\rv{{\textnormal{v}}}
\def\rw{{\textnormal{w}}}
\def\rx{{\textnormal{x}}}
\def\ry{{\textnormal{y}}}
\def\rz{{\textnormal{z}}}

\def\rvepsilon{{\mathbf{\epsilon}}}
\def\rvtheta{{\mathbf{\theta}}}
\def\rva{{\mathbf{a}}}
\def\rvb{{\mathbf{b}}}
\def\rvc{{\mathbf{c}}}
\def\rvd{{\mathbf{d}}}
\def\rve{{\mathbf{e}}}
\def\rvf{{\mathbf{f}}}
\def\rvg{{\mathbf{g}}}
\def\rvh{{\mathbf{h}}}
\def\rvu{{\mathbf{i}}}
\def\rvj{{\mathbf{j}}}
\def\rvk{{\mathbf{k}}}
\def\rvl{{\mathbf{l}}}
\def\rvm{{\mathbf{m}}}
\def\rvn{{\mathbf{n}}}
\def\rvo{{\mathbf{o}}}
\def\rvp{{\mathbf{p}}}
\def\rvq{{\mathbf{q}}}
\def\rvr{{\mathbf{r}}}
\def\rvs{{\mathbf{s}}}
\def\rvt{{\mathbf{t}}}
\def\rvu{{\mathbf{u}}}
\def\rvv{{\mathbf{v}}}
\def\rvw{{\mathbf{w}}}
\def\rvx{{\mathbf{x}}}
\def\rvy{{\mathbf{y}}}
\def\rvz{{\mathbf{z}}}

\def\erva{{\textnormal{a}}}
\def\ervb{{\textnormal{b}}}
\def\ervc{{\textnormal{c}}}
\def\ervd{{\textnormal{d}}}
\def\erve{{\textnormal{e}}}
\def\ervf{{\textnormal{f}}}
\def\ervg{{\textnormal{g}}}
\def\ervh{{\textnormal{h}}}
\def\ervi{{\textnormal{i}}}
\def\ervj{{\textnormal{j}}}
\def\ervk{{\textnormal{k}}}
\def\ervl{{\textnormal{l}}}
\def\ervm{{\textnormal{m}}}
\def\ervn{{\textnormal{n}}}
\def\ervo{{\textnormal{o}}}
\def\ervp{{\textnormal{p}}}
\def\ervq{{\textnormal{q}}}
\def\ervr{{\textnormal{r}}}
\def\ervs{{\textnormal{s}}}
\def\ervt{{\textnormal{t}}}
\def\ervu{{\textnormal{u}}}
\def\ervv{{\textnormal{v}}}
\def\ervw{{\textnormal{w}}}
\def\ervx{{\textnormal{x}}}
\def\ervy{{\textnormal{y}}}
\def\ervz{{\textnormal{z}}}

\def\rmA{{\mathbf{A}}}
\def\rmB{{\mathbf{B}}}
\def\rmC{{\mathbf{C}}}
\def\rmD{{\mathbf{D}}}
\def\rmE{{\mathbf{E}}}
\def\rmF{{\mathbf{F}}}
\def\rmG{{\mathbf{G}}}
\def\rmH{{\mathbf{H}}}
\def\rmI{{\mathbf{I}}}
\def\rmJ{{\mathbf{J}}}
\def\rmK{{\mathbf{K}}}
\def\rmL{{\mathbf{L}}}
\def\rmM{{\mathbf{M}}}
\def\rmN{{\mathbf{N}}}
\def\rmO{{\mathbf{O}}}
\def\rmP{{\mathbf{P}}}
\def\rmQ{{\mathbf{Q}}}
\def\rmR{{\mathbf{R}}}
\def\rmS{{\mathbf{S}}}
\def\rmT{{\mathbf{T}}}
\def\rmU{{\mathbf{U}}}
\def\rmV{{\mathbf{V}}}
\def\rmW{{\mathbf{W}}}
\def\rmX{{\mathbf{X}}}
\def\rmY{{\mathbf{Y}}}
\def\rmZ{{\mathbf{Z}}}

\def\ermA{{\textnormal{A}}}
\def\ermB{{\textnormal{B}}}
\def\ermC{{\textnormal{C}}}
\def\ermD{{\textnormal{D}}}
\def\ermE{{\textnormal{E}}}
\def\ermF{{\textnormal{F}}}
\def\ermG{{\textnormal{G}}}
\def\ermH{{\textnormal{H}}}
\def\ermI{{\textnormal{I}}}
\def\ermJ{{\textnormal{J}}}
\def\ermK{{\textnormal{K}}}
\def\ermL{{\textnormal{L}}}
\def\ermM{{\textnormal{M}}}
\def\ermN{{\textnormal{N}}}
\def\ermO{{\textnormal{O}}}
\def\ermP{{\textnormal{P}}}
\def\ermQ{{\textnormal{Q}}}
\def\ermR{{\textnormal{R}}}
\def\ermS{{\textnormal{S}}}
\def\ermT{{\textnormal{T}}}
\def\ermU{{\textnormal{U}}}
\def\ermV{{\textnormal{V}}}
\def\ermW{{\textnormal{W}}}
\def\ermX{{\textnormal{X}}}
\def\ermY{{\textnormal{Y}}}
\def\ermZ{{\textnormal{Z}}}

\def\vzero{{\bm{0}}}
\def\vone{{\bm{1}}}
\def\vmu{{\bm{\mu}}}
\def\vtheta{{\bm{\theta}}}
\def\va{{\bm{a}}}
\def\vb{{\bm{b}}}
\def\vc{{\bm{c}}}
\def\vd{{\bm{d}}}
\def\ve{{\bm{e}}}
\def\vf{{\bm{f}}}
\def\vg{{\bm{g}}}
\def\vh{{\bm{h}}}
\def\vi{{\bm{i}}}
\def\vj{{\bm{j}}}
\def\vk{{\bm{k}}}
\def\vl{{\bm{l}}}
\def\vm{{\bm{m}}}
\def\vn{{\bm{n}}}
\def\vo{{\bm{o}}}
\def\vp{{\bm{p}}}
\def\vq{{\bm{q}}}
\def\vr{{\bm{r}}}
\def\vs{{\bm{s}}}
\def\vt{{\bm{t}}}
\def\vu{{\bm{u}}}
\def\vv{{\bm{v}}}
\def\vw{{\bm{w}}}
\def\vx{{\bm{x}}}
\def\vy{{\bm{y}}}
\def\vz{{\bm{z}}}

\def\evalpha{{\alpha}}
\def\evbeta{{\beta}}
\def\evepsilon{{\epsilon}}
\def\evlambda{{\lambda}}
\def\evomega{{\omega}}
\def\evmu{{\mu}}
\def\evpsi{{\psi}}
\def\evsigma{{\sigma}}
\def\evtheta{{\theta}}
\def\eva{{a}}
\def\evb{{b}}
\def\evc{{c}}
\def\evd{{d}}
\def\eve{{e}}
\def\evf{{f}}
\def\evg{{g}}
\def\evh{{h}}
\def\evi{{i}}
\def\evj{{j}}
\def\evk{{k}}
\def\evl{{l}}
\def\evm{{m}}
\def\evn{{n}}
\def\evo{{o}}
\def\evp{{p}}
\def\evq{{q}}
\def\evr{{r}}
\def\evs{{s}}
\def\evt{{t}}
\def\evu{{u}}
\def\evv{{v}}
\def\evw{{w}}
\def\evx{{x}}
\def\evy{{y}}
\def\evz{{z}}

\def\mA{{\bm{A}}}
\def\mB{{\bm{B}}}
\def\mC{{\bm{C}}}
\def\mD{{\bm{D}}}
\def\mE{{\bm{E}}}
\def\mF{{\bm{F}}}
\def\mG{{\bm{G}}}
\def\mH{{\bm{H}}}
\def\mI{{\bm{I}}}
\def\mJ{{\bm{J}}}
\def\mK{{\bm{K}}}
\def\mL{{\bm{L}}}
\def\mM{{\bm{M}}}
\def\mN{{\bm{N}}}
\def\mO{{\bm{O}}}
\def\mP{{\bm{P}}}
\def\mQ{{\bm{Q}}}
\def\mR{{\bm{R}}}
\def\mS{{\bm{S}}}
\def\mT{{\bm{T}}}
\def\mU{{\bm{U}}}
\def\mV{{\bm{V}}}
\def\mW{{\bm{W}}}
\def\mX{{\bm{X}}}
\def\mY{{\bm{Y}}}
\def\mZ{{\bm{Z}}}
\def\mBeta{{\bm{\beta}}}
\def\mPhi{{\bm{\Phi}}}
\def\mLambda{{\bm{\Lambda}}}
\def\mSigma{{\bm{\Sigma}}}

\newcommand{\tens}[1]{\bm{\mathsfit{#1}}}
\def\tA{{\tens{A}}}
\def\tB{{\tens{B}}}
\def\tC{{\tens{C}}}
\def\tD{{\tens{D}}}
\def\tE{{\tens{E}}}
\def\tF{{\tens{F}}}
\def\tG{{\tens{G}}}
\def\tH{{\tens{H}}}
\def\tI{{\tens{I}}}
\def\tJ{{\tens{J}}}
\def\tK{{\tens{K}}}
\def\tL{{\tens{L}}}
\def\tM{{\tens{M}}}
\def\tN{{\tens{N}}}
\def\tO{{\tens{O}}}
\def\tP{{\tens{P}}}
\def\tQ{{\tens{Q}}}
\def\tR{{\tens{R}}}
\def\tS{{\tens{S}}}
\def\tT{{\tens{T}}}
\def\tU{{\tens{U}}}
\def\tV{{\tens{V}}}
\def\tW{{\tens{W}}}
\def\tX{{\tens{X}}}
\def\tY{{\tens{Y}}}
\def\tZ{{\tens{Z}}}

\def\gA{{\mathcal{A}}}
\def\gB{{\mathcal{B}}}
\def\gC{{\mathcal{C}}}
\def\gD{{\mathcal{D}}}
\def\gE{{\mathcal{E}}}
\def\gF{{\mathcal{F}}}
\def\gG{{\mathcal{G}}}
\def\gH{{\mathcal{H}}}
\def\gI{{\mathcal{I}}}
\def\gJ{{\mathcal{J}}}
\def\gK{{\mathcal{K}}}
\def\gL{{\mathcal{L}}}
\def\gM{{\mathcal{M}}}
\def\gN{{\mathcal{N}}}
\def\gO{{\mathcal{O}}}
\def\gP{{\mathcal{P}}}
\def\gQ{{\mathcal{Q}}}
\def\gR{{\mathcal{R}}}
\def\gS{{\mathcal{S}}}
\def\gT{{\mathcal{T}}}
\def\gU{{\mathcal{U}}}
\def\gV{{\mathcal{V}}}
\def\gW{{\mathcal{W}}}
\def\gX{{\mathcal{X}}}
\def\gY{{\mathcal{Y}}}
\def\gZ{{\mathcal{Z}}}

\def\sA{{\mathbb{A}}}
\def\sB{{\mathbb{B}}}
\def\sC{{\mathbb{C}}}
\def\sD{{\mathbb{D}}}
\def\sF{{\mathbb{F}}}
\def\sG{{\mathbb{G}}}
\def\sH{{\mathbb{H}}}
\def\sI{{\mathbb{I}}}
\def\sJ{{\mathbb{J}}}
\def\sK{{\mathbb{K}}}
\def\sL{{\mathbb{L}}}
\def\sM{{\mathbb{M}}}
\def\sN{{\mathbb{N}}}
\def\sO{{\mathbb{O}}}
\def\sP{{\mathbb{P}}}
\def\sQ{{\mathbb{Q}}}
\def\sR{{\mathbb{R}}}
\def\sS{{\mathbb{S}}}
\def\sT{{\mathbb{T}}}
\def\sU{{\mathbb{U}}}
\def\sV{{\mathbb{V}}}
\def\sW{{\mathbb{W}}}
\def\sX{{\mathbb{X}}}
\def\sY{{\mathbb{Y}}}
\def\sZ{{\mathbb{Z}}}

\def\emLambda{{\Lambda}}
\def\emA{{A}}
\def\emB{{B}}
\def\emC{{C}}
\def\emD{{D}}
\def\emE{{E}}
\def\emF{{F}}
\def\emG{{G}}
\def\emH{{H}}
\def\emI{{I}}
\def\emJ{{J}}
\def\emK{{K}}
\def\emL{{L}}
\def\emM{{M}}
\def\emN{{N}}
\def\emO{{O}}
\def\emP{{P}}
\def\emQ{{Q}}
\def\emR{{R}}
\def\emS{{S}}
\def\emT{{T}}
\def\emU{{U}}
\def\emV{{V}}
\def\emW{{W}}
\def\emX{{X}}
\def\emY{{Y}}
\def\emZ{{Z}}
\def\emSigma{{\Sigma}}

\newcommand{\etens}[1]{\mathsfit{#1}}
\def\etLambda{{\etens{\Lambda}}}
\def\etA{{\etens{A}}}
\def\etB{{\etens{B}}}
\def\etC{{\etens{C}}}
\def\etD{{\etens{D}}}
\def\etE{{\etens{E}}}
\def\etF{{\etens{F}}}
\def\etG{{\etens{G}}}
\def\etH{{\etens{H}}}
\def\etI{{\etens{I}}}
\def\etJ{{\etens{J}}}
\def\etK{{\etens{K}}}
\def\etL{{\etens{L}}}
\def\etM{{\etens{M}}}
\def\etN{{\etens{N}}}
\def\etO{{\etens{O}}}
\def\etP{{\etens{P}}}
\def\etQ{{\etens{Q}}}
\def\etR{{\etens{R}}}
\def\etS{{\etens{S}}}
\def\etT{{\etens{T}}}
\def\etU{{\etens{U}}}
\def\etV{{\etens{V}}}
\def\etW{{\etens{W}}}
\def\etX{{\etens{X}}}
\def\etY{{\etens{Y}}}
\def\etZ{{\etens{Z}}}

\newcommand{\pdata}{p_{\rm{data}}}
\newcommand{\ptrain}{\hat{p}_{\rm{data}}}
\newcommand{\Ptrain}{\hat{P}_{\rm{data}}}
\newcommand{\pmodel}{p_{\rm{model}}}
\newcommand{\Pmodel}{P_{\rm{model}}}
\newcommand{\ptildemodel}{\tilde{p}_{\rm{model}}}
\newcommand{\pencode}{p_{\rm{encoder}}}
\newcommand{\pdecode}{p_{\rm{decoder}}}
\newcommand{\precons}{p_{\rm{reconstruct}}}

\newcommand{\laplace}{\mathrm{Laplace}} 

\newcommand{\E}{\mathbb{E}}
\newcommand{\Ls}{\mathcal{L}}
\newcommand{\R}{\mathbb{R}}
\newcommand{\emp}{\tilde{p}}
\newcommand{\lr}{\alpha}
\newcommand{\reg}{\lambda}
\newcommand{\rect}{\mathrm{rectifier}}
\newcommand{\softmax}{\mathrm{softmax}}
\newcommand{\sigmoid}{\sigma}
\newcommand{\softplus}{\zeta}
\newcommand{\KL}{D_{\mathrm{KL}}}
\newcommand{\Var}{\mathrm{Var}}
\newcommand{\standarderror}{\mathrm{SE}}
\newcommand{\Cov}{\mathrm{Cov}}
\newcommand{\normlzero}{L^0}
\newcommand{\normlone}{L^1}
\newcommand{\normltwo}{L^2}
\newcommand{\normlp}{L^p}
\newcommand{\normmax}{L^\infty}

\newcommand{\parents}{Pa} 

\let\ab\allowbreak

%% file: section/1_intro.tex
\vspace{-20pt}
\section{Introduction}
\label{sec:intro}
\vspace{-5pt}

\begin{figure}[t]
    \centering
    \includegraphics[width=0.8\linewidth]{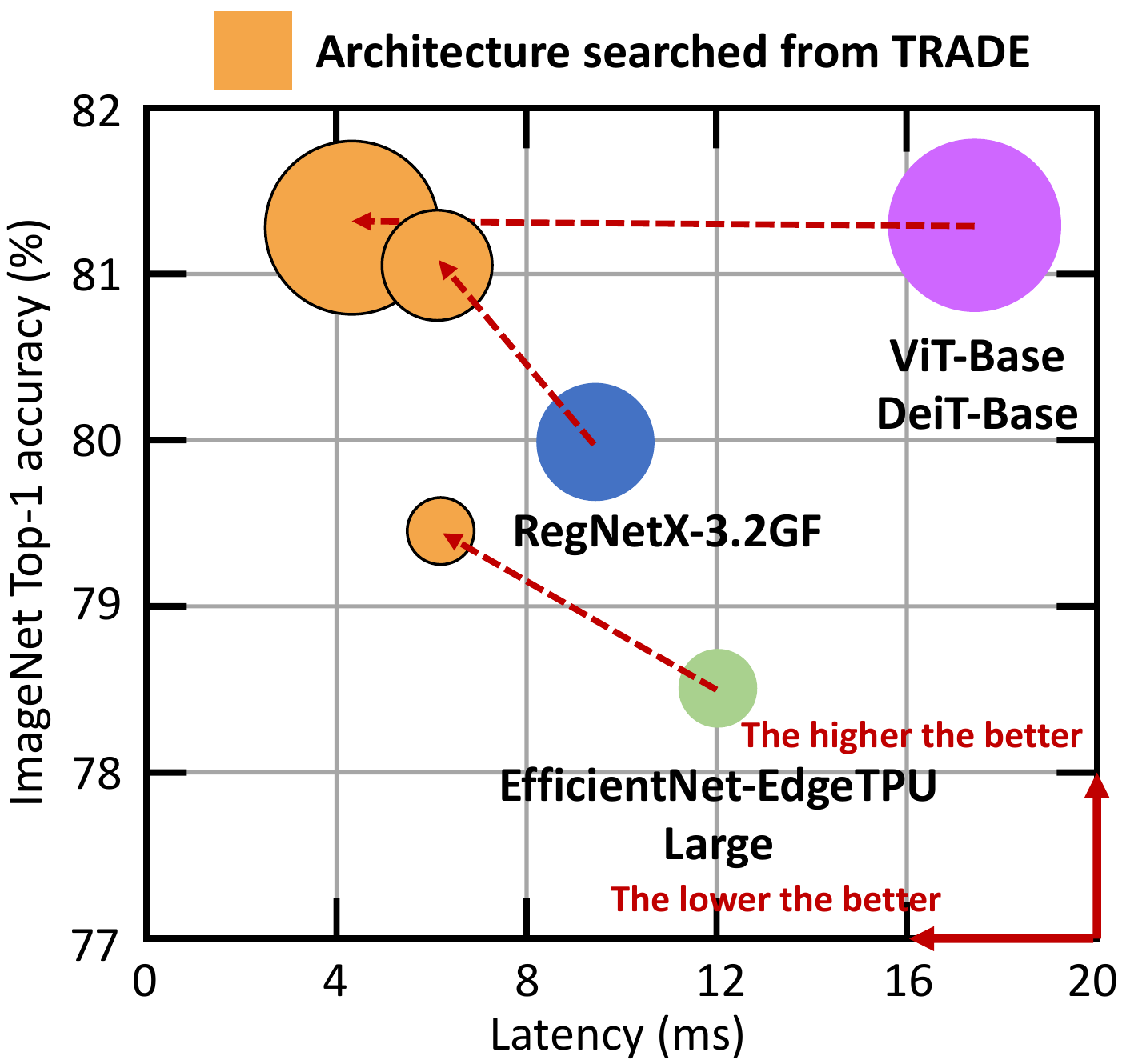}
    \vspace{-10pt}
    \caption{Comparison between TRADE and recent state-of-the-art models on ImageNet when KD is implemented on all models. Orange circles indicate the architectures obtained from TRADE while each of them shares the same teachers as the pre-defined architecture being compared. TRADE provides further optimal architecture family for knowledge distillation. Here, used teachers are in \autoref{tab:tea_stu_acc} when using 224-pixel images.}
    \label{fig:overview_acc}
    \vspace{-10pt}
\end{figure}
\begin{figure}[t]
    \centering
    \includegraphics[width=0.8\linewidth]{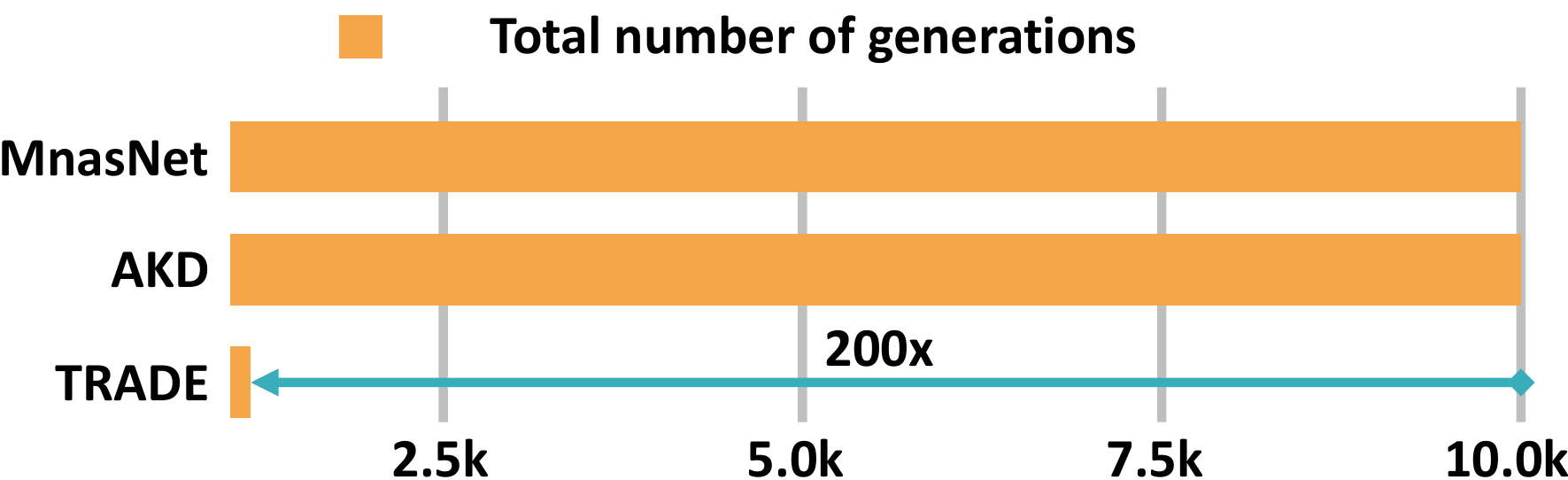}
    \vspace{-10pt}
    \caption{TRADE reduces design costs by orders of magnitude compared to other NAS methods\,\cite{tan2019mnasnet, liu2020search}.}
    \label{fig:overview_cost}
    \vspace{-15pt}
\end{figure}
\begin{figure*}[t]
    \centering
    \includegraphics[width=0.9\linewidth]{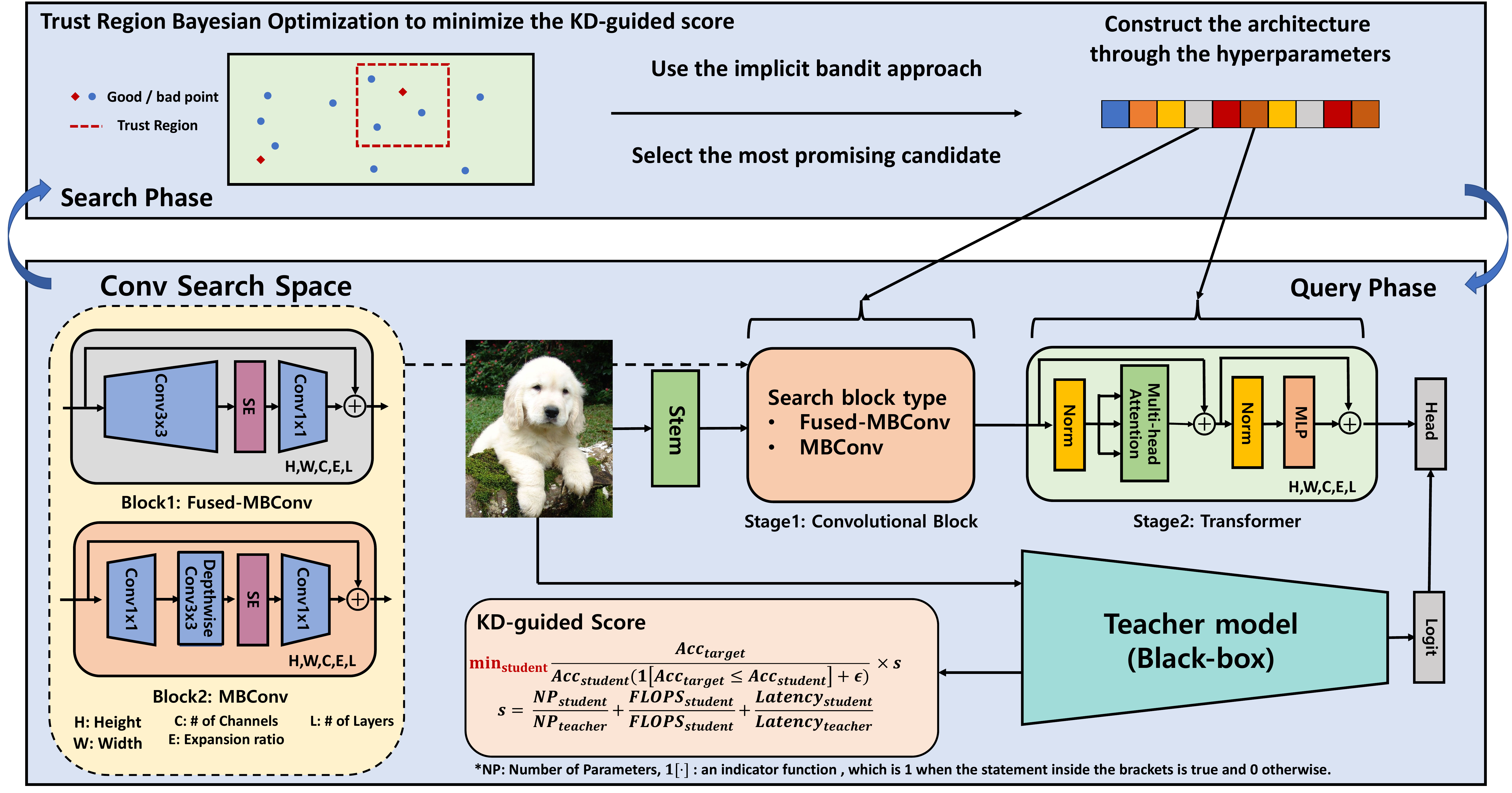}\vspace{-5pt}

    \caption{Overview of the proposed TRADE search algorithm. It shows our overall NAS flow, consisting of three major components: a BO for architecture search, a search space for multiple operators, and a training environment including KD. In the search phase, we jointly search an optimal combination of recent convolution blocks and the Transformer operations for the design of the student model using trust region BO by minimizing the KD-guided score referred to \cite{eriksson2019scalable, pmlr-v123-kim20a, moons2021distilling}}
    \label{fig:overview}
    \vspace{-15pt}
\end{figure*}

Convolutional Neural Networks\,(CNN) have achieved massive success in many computer vision tasks, including image classification\,\cite{szegedy2016rethinking, DBLP:journals/corr/HanKK16, DBLP:journals/corr/ZagoruykoK16, DBLP:journals/corr/abs-1811-06965}, object detection\,\cite{DBLP:journals/corr/LiuAESR15}, and semantic segmentation\,\cite{DBLP:journals/corr/ChenPK0Y16}. Recently, Vision Transformer\,(ViT) has shown that with mostly vanilla Transformer layers, one could obtain reasonable performance on ImageNet-1K alone\,\cite{dosovitskiy2020image}. Following this direction, a variety of architectures appears\,\cite{touvron2021training, tolstikhin2021mlp}. For the purpose of obtaining more precise results, such models frequently develop into deeper and broader structures\,\cite{tan2019efficientnet, sandler2018mobilenetv2, DBLP:journals/corr/ZagoruykoK16, DBLP:journals/corr/HuangLW16a, dosovitskiy2020image, tan2021efficientnetv2}\,(\autoref{fig:overview_acc}). On the other hand, there is an increasing demand for deploying state-of-the-art models under restricted resource budgets such as the number of parameters and FLOPS. For example, some devices, unfortunately, have limited computational resources, and thus, they are not able to run state-of-the-art algorithms.

Knowledge Distillation\,(KD) is one of the most potent model compression solutions by transferring knowledge from a cumbersome model to a small model\,\cite{hinton2015distilling}. Most existing works related to KD focus on improving generalization ability by optimizing the neural network's output distribution\,\cite{romero2014fitnets, zagoruyko2016paying, srinivas2018knowledge, kim2018paraphrasing, heo2019knowledge, heo2019comprehensive, Zhang_2019_ICCV, park2019relational, DBLP:journals/corr/abs-2006-05525, DBLP:journals/corr/abs-2105-08919}. Meanwhile, the best student designs for various teacher models trained on the same task and dataset may differ from one another.\,\cite{liu2020search}. To alleviate this issue, \textit{Liu et. al}\,\cite{liu2020search} propose a new generalized approach for KD, referred to as Architecture-aware Knowledge Distillation\,(AKD), which finds the best student architectures for distilling the given teacher model. The authors\,\cite{liu2020search} use Reinforcement Learning\,(RL) to automatically design neural architectures with a KD-based reward function. However, AKD not only requires lots of GPU hours even in a reduced setting\,(\autoref{fig:overview_cost}), but it lacks on understanding the structural knowledge when recent architectures are used in KD.
Furthermore, recent architecture-aware KD studies\,\cite{liu2020search, moons2021distilling, kang2020towards} have not considered the search space including recent convolutional blocks and the Transformer simultaneously, yet.




In this paper, we propose an efficient search method of the optimal architecture for knowledge distillation, called \textit{\underline{T}rust \underline{R}egion \underline{A}ware architecture search to \underline{D}istill knowledge \underline{E}ffectively}\,(TRADE), motivated by Bayesian AutoML works\,\cite{DBLP:journals/corr/abs-2106-11890, eriksson2019scalable, shi2019bridging, diouane2021trego}. Our key idea is to utilize the trust region BO to allocate samples across multiple local regions and thus pick which local optimizations runs to continue at each iteration by exploiting an implicit multi-armed bandit approach. TRADE provides highly sample-efficient optimization of multiple competing objectives; (1)\,Number of Parameters\,(NP), (2)\,FLOPS, (3)\,latency, and (4)\,accuracy. TRADE aims to minimize the NP, FLOPS, and latency while maximizing the accuracy\,(\autoref{fig:overview}). The contributions of this paper are as follows:
\vspace{-10pt}
\begin{itemize}
    \item We propose an efficient framework for a quick architecture search under any teacher model by combining multiple modules\,(e.g., MBConv\,\cite{sandler2018mobilenetv2}, Fused-MBConv\,\cite{tan2021efficientnetv2}, and the Transformer\,\cite{dosovitskiy2020image}). Specifically, referring to \cite{xiao2021early, dai2021coatnet}, we design the search space to allow the convolutional layer to take charge of early visual processing and leave the representation learning to the Transformer\,(\autoref{fig:overview}). This approach imposes an extreme search efficiency under the variety of teacher models with consideration, allowing an extensible search space\,(\autoref{fig:overview_cost}).
    
    \item We empirically validate that TRADE finds a more accurate and efficient student model than conventional pre-defined architectures under KD training by minimizing our new KD-guided score containing the NP, FLOPS, latency, and top1 accuracy\,(\autoref{fig:overview_acc}, \autoref{fig:overview}). 
    
\end{itemize}
\vspace{-10pt}


%% file: section/2_preliminary.tex
\section{Related work}
\label{sec:prelim}

\vspace{-5pt}
\subsection{Knowledge Distillation}
Let us denote the softened probability vector in a network $f$ as $p^{f}(\tau)$ where $\tau$ is a temperature scaling hyperparameter. The $k$-th element in $p^{f}(\tau)$ is defined as $p^{f}(\tau)_k = \frac{e^{z^f_k/\tau}}{\sum_j e^{z^f_j/\tau}}$ where $z^f$ is a logit vector which is an input to the softmax function and $z^f_k$ is the $k$-th element in $z^f$. Then, the typical loss for a student network is a linear combination of the cross entropy (CE) loss $\mathcal{L}_{CE}$ and the KD loss $\mathcal{L}_{KD}$:
\vspace{-5pt}
\begin{equation}\label{equation1}\small
\begin{gathered} 
    \mathcal{L} = (1-\alpha) \mathcal{L}_{CE} + \alpha \mathcal{L}_{KD}, \\
    \text{where} \quad \mathcal{L}_{KD} = \tau^2  \sum_{j} p^{t}(\tau)_j \log \frac{p^{t}(\tau)_j}{p^{s}(\tau)_j}
\end{gathered}
\end{equation}
where $s$ is a student network, $t$ is a teacher network, $\mathcal{L}_{CE}$ is the loss between the student prediction and ground-truth label, and $\alpha$ is a hyperparameter of the linear combination. 
\vspace{-10pt}
\subsection{Sample-based Neural Architecture Search}

Sample-based NAS first searches a number of viable candidate architectures and then evaluates their performances. Early sample-based methods use RL\,\cite{tan2019mnasnet,zoph2016neural} or ES\,\cite{liu2017hierarchical,real2019regularized,real2017large}, which requires heavy computation for sampling a considerable amount of candidates. Bayesian optimization\,(BO) is being a key solution of sample-efficient NAS framework by leveraging a probabilistic surrogate model, typically a Gaussian process\,(GP)\,\cite{shi2019bridging,white2019bananas,kandasamy2018neural,wang2021sample,DBLP:journals/corr/abs-2106-11890}. The rising prevalence of mobile environments warrants a need to balance accuracy performance and resources\,(e.g., inference time, FLOPS). Multi-Objective Optimization\,(MOO)\,\cite{eriksson2021latency} can address this issue. The goal is to maximize/minimize a vector-valued objective $\mathbf{f}(\mathbf{x})\in\mathbb{R}^{d_x}$ over a bounded set $\sX\subset\mathbb{R}^d$ while seeking the Pareto frontier: the set of solutions where improving one objective will degrade another. TRADE rapidly finds the Pareto frontier by minimizing our new KD-guided score constrained with teacher specifications, as we will see in the experiments.

%% file: section/3_method.tex
\vspace{-10pt}
\section{Method: TRADE} \label{sec:method}
\vspace{-5pt}
In this section, we study the efficacy of the search algorithm and the OR and introduce our TRADE framework with our new KD-guided score and search space.





\begin{figure}[t]
\centering
    \includegraphics[width=1.0\linewidth]{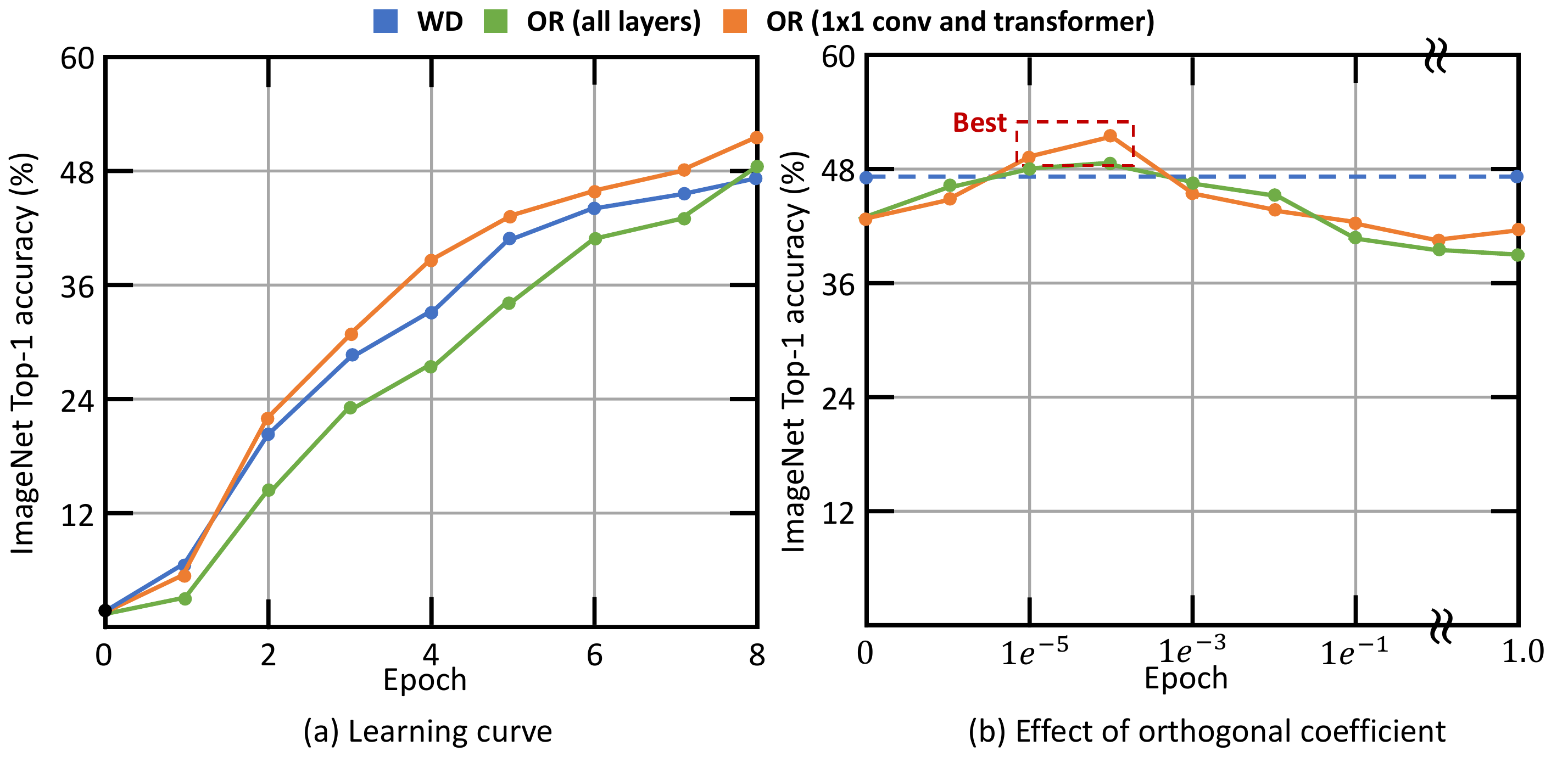}
    \vspace{-20pt}
    \caption{Efficacy of applying the OR in 1x1 convolution and Transformer operations. (a): learning curve of ViT-Large-32-R50; (b) effect of orthogonal coefficient in training ViT-Large-32-R50} \label{fig:pre_ortho}
    \vspace{-15pt}
\end{figure}

\vspace{-5pt}
\paragraph{Efficiency on 1x1 convolution and Transformer.} 1x1 convolutional operators and FFN layers in the Transformer can bring the disadvantage that the number of total parameters can grow very high. Therefore, it is necessary to reduce redundancy in such high dimensions. One of the most potent training tricks is to apply an OR to the weight matrix\,\cite{DBLP:journals/corr/abs-1810-09102,pmlr-v123-kim20a,9804718}. We empirically observe that making 1x1 convolutions and FFN layers in the Transformer block to be near semi-orthogonal works well in the proxy task but also facilitates generalization\,(\autoref{fig:pre_ortho}). This is an extension of the previous findings\,\cite{pmlr-v123-kim20a}. On the other hand, the OR cost function to either convolutional operators with kernel size 3 or fully-connected layer seems to hinder finding a good filter to extract spatial information rather than to fix the vanishing/exploding gradient issue. Even recent direct OR constraint also has difficulty in learning the isometric property\,\cite{wang2020orthogonal,qi2020deep}. From this result, in Section \ref{sec4:ex}, we train the searched network by regularizing OR only on 1x1 convolutions and FFN layers in the Transformer block. In addition, we also apply the same strategy for the search phase on a proxy task to obtain the reward.

\begin{table}[t]
\centering\footnotesize
\caption{Various block types used in neural architecture methods.\label{tab:block-type}} 
\begin{tabular}{c|cccc}
\toprule
Algorithm          & MBConv & Fused-MBConv & Transformer  \\ \midrule\midrule
AKD                & \checkmark &              &                  \\
DONNA                & \checkmark &              &                  \\
EfficientNetV2     & \checkmark & \checkmark &                  \\
ViT               &         &              & \checkmark      \\
CoATNet         & \checkmark &              & \checkmark \\
\rowcolor{LightCyan}TRADE\,(Ours)              & \checkmark & \checkmark & \checkmark   \\\bottomrule
\end{tabular}
\vspace{-20pt}
\end{table}

\vspace{-10pt}
\subsection{KD-guided Bayesian NAS}

Our TRADE leverages TurBO\,\cite{eriksson2019scalable} for NAS to rapidly explore student architectures that are comparable with existing architectures. In this section, we introduce our multiple design choices and discuss our insights. In Multi-Objective Optimization\,(MOO), our target is to maximize the model accuracy as well as to minimize the NP, FLOPS, and latency.


\begin{table*}[t]
\vspace{-5pt}
\centering\footnotesize\addtolength{\tabcolsep}{-1.0pt}
\caption{Architectures searched by TRADE performance results on ImageNet. TRADE outperforms the performance of other pre-defined models with respect to NP, FLOPS, latency and top1 accuracy. HL indicates the hard label, and D indications the distillation.\label{tab:tea_stu_acc}} 
\begin{tabular}{ccccccccc}
\toprule
Teacher                  & Student  & FLOPS &  NP & Latency & searching by & training by  & top-1 & top-5 \\\midrule \midrule
\multirow{8}{*}{RegNetY-8.0GF}          & \multirow{2.5}{*}{RegNetX-3.2GF}         &  \multirow{2.5}{*}{3.2 B}  & \multirow{2.5}{*}{15.3 M}  & \multirow{2.5}{*}{8.5 ms}                  & -            & HL   &  80.3     &  93.1  \\\cmidrule{6-9}
                                        &                                        &                          &                          &                                          & -            & D &  80.9     &  93.9  \\\cmidrule{2-9}
                                        & \multirow{2.5}{*}{TRADE-Reg$^\dagger$ w/o KD}             &  \multirow{2.5}{*}{{1.6 B}}  &  \multirow{2.5}{*}{{15.7 M}}  & \multirow{2.5}{*}{{6.2 ms}}                & HL   & HL   &  80.5     &  93.1  \\\cmidrule{6-9}
                                        &                                        &                          &                           &                                        & HL   & D &  80.9     &  93.9  \\\cmidrule{2-9}
                                        &\multirow{2.5}{*}{TRADE-Reg}             & \multirow{2.5}{*}{\textbf{1.5 B}}  &  \multirow{2.5}{*}{\textbf{14.8 M}}  & \multirow{2.5}{*}{\textbf{6.0 ms}} & HL   &  D & 80.8     &  93.6  \\\cmidrule{6-9}
\rowcolor{LightCyan}\cellcolor{white}   & \cellcolor{white} & \cellcolor{white} & \cellcolor{white} & \cellcolor{white} &  \textbf{D} & \textbf{D} &  \textbf{81.0} &  \textbf{94.1}    \\\midrule
\multirow{8}{*}{ViT-Large-16}      & ViT-Small-Patch16      &  4.2 B &  22.1 M      & 18.1 ms      & -            & HL   &  81.4     &  96.1  \\\cmidrule{2-9}
                                        & DeiT-Small-Patch16 & 4.2 B & 22.4 M & 18.5 ms & -            & D &  81.2     &  95.4  \\\cmidrule{2-9}
                                        & \multirow{2.5}{*}{TRADE-ViT$^\dagger$ w/o KD}             & \multirow{2.5}{*}{{2.1 B}}  & \multirow{2.5}{*}{{23.2 M}}                 &  \multirow{2.5}{*}{{5.6 ms}}      & HL   & HL   &  81.2     &  95.2  \\\cmidrule{6-9}
                                        &                                        &   &                                        &                               & HL   & D &  81.1   & 95.5 \\\cmidrule{2-9}
                                        & \multirow{2.5}{*}{TRADE-ViT}             & \multirow{2.5}{*}{\textbf{1.8 B}}  & \multirow{2.5}{*}{\textbf{21.4 M}}                 &  \multirow{2.5}{*}{\textbf{4.5 ms}}        & D & HL   &  80.9  & 95.9 \\\cmidrule{6-9}
\rowcolor{LightCyan}\cellcolor{white}   & \cellcolor{white} & \cellcolor{white} & \cellcolor{white} & \cellcolor{white} & \textbf{D} & \textbf{D} &  \textbf{81.5} &  \textbf{95.9}    \\\midrule
\multirow{11}{*}{EfficientNetV2-M}      & \multirow{2}{*}{EfficientNet-EdgeTPU-Large}      & \multirow{2.5}{*}{4.0 B}  &  \multirow{2.5}{*}{10.6 M}               & \multirow{2.5}{*}{12.8 ms}   & -            & HL   &  77.2     &  93.6  \\\cmidrule{6-9}
                                        &                                        &   &                                        &                           & -            & D &  78.4     &  93.5  \\\cmidrule{2-9}
                                        & \multirow{2.5}{*}{EfficientNetV2-S}      & \multirow{2.5}{*}{2.7 B}  &  \multirow{2.5}{*}{21.5 M}               & \multirow{2.5}{*}{\textbf{5.4 ms}}   & -            & HL   &  80.2     &  94.9  \\\cmidrule{6-9}
                                        &                                        &   &                                        &                           & -            & D &  80.1     &  94.7  \\\cmidrule{2-9}

                                        & \multirow{2.5}{*}{TRADE-EffNet$^\dagger$ w/o KD}          & \multirow{2.5}{*}{{2.7 B}}  & \multirow{2.5}{*}{{10.9 M}}                & \multirow{2.5}{*}{7.1 ms}   & HL   & HL   &  80.1     &  95.0  \\\cmidrule{6-9}
                                        &                                        &   &                                        &                           & HL   & D & 79.9 & 94.9 \\\cmidrule{2-9}
                                        & \multirow{2.5}{*}{TRADE-EffNet}          & \multirow{2.5}{*}{\textbf{2.1 B}}  & \multirow{2.5}{*}{\textbf{8.8 M}}                & \multirow{2.5}{*}{6.0 ms}   & D & HL   & 79.5 & 93.9 \\\cmidrule{6-9}
\rowcolor{LightCyan}\cellcolor{white}   & \cellcolor{white} & \cellcolor{white} & \cellcolor{white} & \cellcolor{white} &\textbf{D} & \textbf{D} &  \textbf{80.3} &  \textbf{95.0}    \\\bottomrule
\end{tabular}
\vspace{-15pt}
\end{table*}



\vspace{-10pt}
\paragraph{KD-guided Score.}
TRADE wishes to find the optimal architecture which minimizes the score referred to as the evaluation metric of NeurIPS MicroNet Challenge 2019\,\cite{pmlr-v123-kim20a}:
\begin{equation}
\begin{gathered}
   \text{Score} = \frac{\text{Acc}_{target}}{\text{Acc}_{student}({\mathbf{1}[\text{Acc}_{target}\leq \text{Acc}_{student}]+\epsilon)} } \times s \\
    s = \frac{\text{NP}_{student}}{\text{NP}_{teacher}} +  \frac{\text{FLOPS}_{student}}{\text{FLOPS}_{teacher}} + \frac{\text{Latency}_{student}}{\text{Latency}_{teacher}}
\end{gathered}\label{eq:reward}
\end{equation}

\noindent where $\mathbf{1}[\cdot]$ is the indicator function, which is 1 when the statement inside the brackets is true and 0 otherwise, and Acc$_{model}$ denotes the top1 accuracy of the model. We measure the latency on the NVIDIA 2080-Ti GPU. The TRADE algorithm creates an architecture with a minimum target accuracy while becoming lighter as the value of $s$ is lower. Advances from prior works\,\cite{tan2019mnasnet,liu2020search,moons2021distilling}, this score designs candidates more specialized for distillation in that it compresses and considers the teacher's resource budget at the same time.


\vspace{-10pt}




%% file: section/4_result.tex
\section{Results}\label{sec4:ex}

\vspace{-5pt}
\paragraph{Performance improvements.} \autoref{tab:tea_stu_acc} shows the prominence of all searched models from our TRADE framework. The series of networks from TRADE generally use an order of magnitude, fewer parameters and FLOPS, and lower latency than other pre-defined networks while having higher accuracy. These gains come from both better architecture and better regularization tricks. \autoref{fig:overview_acc}\,(a) illustrates the Latency-Accuracy curve where our TRADE facilitates the KD performance and dramatically reduces the search cost\,(e.g., search iterations) compared to RL-based methods\,\cite{liu2020search,tan2019mnasnet}\,(\autoref{fig:overview_cost}). \autoref{fig:search_efficacy} shows the efficacy of TRADE on the number of samples when we replace the trust region BO with other search methods such as random and LA-MCTS\,\cite{wang2020learning}. An OR is also used in both search methods (Appendix).




\vspace{-10pt}
\paragraph{Percentage of training samples.}
In practice, using all training examples may bring a significant burden during the search process. Thus, to reflect the practical scenario of model search, we conduct the experiments by varying the percentage of training examples from 10\% to 50\% while keeping the number of iterations. For instance, if 10\% of training data is used, then the model is trained with 90 epochs; otherwise, the model is trained with 9 epochs under the whole training data. \autoref{tab:proxy-training} shows utilizing the portion of training samples also keep the order of architecture rank when keeping the number of iterations under KD scenarios. The candidates in \autoref{tab:proxy-training} are selected from the Pareto-frontier.

\begin{table}[t]
\vspace{-5pt}
\centering \small\addtolength{\tabcolsep}{-3.5pt}
\centering\caption{All student models are conducted under the supervision of the ViT-Large-16 model.\label{tab:proxy-training}}
\begin{tabular}{c|cccccc|c} \toprule
\multirow{2.5}{*}{Candidates}& \multicolumn{6}{c|}{Proxy task}           & \multirow{2.5}{*}{Original} \\\cmidrule{2-7}
         & 10\% & 20\% & 30\% & 40\% & 50\% & \multicolumn{1}{c|}{100\%} &                           \\ \midrule\midrule
Pareto 1 & 40.6 & 42.0 & 43.5 & 44.8 & 45.1 & 48.7 & 73.3 \\ 
Pareto 2 & 42.1 & 44.5 & 45.3 & 46.7 & 47.0 & 51.2 & 76.2 \\ 
Pareto 3 & 48.7 & 49.1 & 49.5 & 51.2 & 52.5 & 55.4 & 81.5 \\ \bottomrule
\end{tabular}
  \vspace{-20pt}
\end{table}

%% file: section/5_conclusion.tex
\vspace{-10pt}
\section{Conclusion} \label{sec:concl}
\vspace{-5pt}
This paper introduces an efficient algorithm, called TRADE, for automatically searching the optimal architecture for distillation by using the trust region Bayesian optimization. 
We show that the searched networks from TRADE are more superior to other existing pre-defined architectures under KD situations in respect of NP, FLOPS, latency, and top1 accuracy. 
We believe that TRADE opens the door to rapidly search the most structure-similar student model.

%% file: appendix/app1.tex
\section{Related Works}
\subsection{Computational Costs in Neural Architecture Search}
Neural Architecture Search\,(NAS) studies have also suffered from such computing time issues. Recent NAS studies have been involved in leveraging RL\,\cite{zoph2016neural, tan2019mnasnet}, Evolutionary Search\,(ES)\,\cite{real2019regularized, liu2017hierarchical, real2017large}, and weight-sharing\,\cite{cai2019once, yu2020bignas, wang2021attentivenas}, among others. However, RL and ES can incur prohibitively high computational costs because they require training and evaluating a large number of architectures. While weight sharing can improve sample complexity, the performance of candidate architectures during the search phase does not represent their true performance in the evaluation phase\,\cite{pham2018efficient, shi2019bridging, yu2019evaluating, zhao2021few}. Bayesian optimization\,(BO) may struggle with limited samples and high dimensional space, but still has led to recent success in NAS\,\cite{white2019bananas, kandasamy2018neural, parsa2020bayesian, wang2021sample, DBLP:journals/corr/abs-2106-11890} as well as tuning machine learning hyperparameters\,\cite{DBLP:journals/corr/abs-2104-10201}.  

Finding an appropriate architecture for a neural network to solve a deep learning problem is tedious and difficult, requiring a lot of expert knowledge and heuristics.  To automate this task, NAS methods have been evolving mainly in two directions: (1) one-shot NAS and (2) sample-based NAS. In general, an one-shot NAS encodes the entire search space into a single supernet using weight sharing\,\cite{pham2018efficient,cai2019once,wang2021attentivenas,yu2020bignas} or continuous relaxation\,\cite{liu2018darts}, but it suffers from initialization sensitivity\,\cite{liu2018darts,shi2019bridging}, ranking correlation issues\,\cite{chu2021scarlet}, and memory bank issues\,\cite{shi2019bridging}.

\subsection{Neural Architecture for Computer Vision}


With the rapid growth of interest in deep neural networks, recent neural architectures have evolved into two folds: (1) factorized convolutions which factorize a standard convolution into a depthwise convolution and a 1×1 convolution\,\cite{howard2017mobilenets,sandler2018mobilenetv2}, and (2) self-attention to feature activation map. Factorized convolutions serve as a solution for running powerful CNNs on devices with limited computational resources. Tan et al.\,\cite{tan2019efficientnet,tan2021efficientnetv2} propose a simple scaling method for better generalization and EfficientNet families having parameter efficiency as well as high performance. Radosavovic et al.\,\cite{radosavovic2020designing} use network design spaces that parameterize populations of networks instead of using the aforementioned classic design spaces. On the other hand, Dosovitskiy et al.\,\cite{dosovitskiy2020image} incorporate self-attention to tackle image recognition with no convolution by presenting the Transformer as a solution, called ViT. CoAtNet\,\cite{dai2021coatnet} is presented as another solution by combining convolution and self-attention into a single computational block and vertically stacking convolution layers and attention layers. However, there still remains a lack in the study of combining recent promising blocks in one architecture due to its challenges of requiring a much larger design space and much more expensive tuning cost. Under the KD framework, TRADE can disentangle such issues with rapid Bayesian search.

\subsection{Orthogonality Regularization\,(OR)}
An OR is a class of regularization methods that makes the weight matrices of a model almost semi-orthogonal. Applying an OR to many layers could accelerate the learning since the OR can solve the vanishing/exploding gradient issue through the norm preservation property\,\cite{DBLP:journals/corr/XieXP17, 9804718}. In a nutshell, consider a vector $\mathbf{x} \in \mathbb{R}^{d_\mathbf{x}}$ that is mapped by a linear transformation $\mathbf{W} \in \mathbb{R}^{d_{\mathbf{x}}\times d_{\mathbf{y}}}$ to another vector $\mathbf{y} \in \mathbb{R}^{d_\mathbf{y}}$\,(i.e., $\mathbf{y} = \mathbf{W}^\top\mathbf{x}$). Then, the below satisfies when a weight matrix is semi-orthogonal\,(i.e., $\mathbf{W}\mathbf{W}^\top=\mathbf{I}_{d_{\mathbf{y}}}$; $\mathbf{I}_{d_{\mathbf{y}}}$ is an identity matrix).
\vspace{-5pt}
\begin{equation}
    \|\mathbf{y}\| = \sqrt{\mathbf{y}^\top \mathbf{y}} = \sqrt{\mathbf{x}^\top \mathbf{W} \mathbf{W}^\top \mathbf{x}} = \sqrt{\mathbf{x}^\top \mathbf{x}} = \| \mathbf{x} \|
\end{equation}
\vspace{-10pt}

\noindent where $\|\cdot\|$ denotes the Euclidean norm. To apply an OR into convolutional operators, it is generally necessary to reshape each convolutional operator into a matrix  $\mathbf{W} \in \mathbb{R}^{m \times n}$ with $m =C_{out} $ and $n =S \times H \times C_{in}$, where $S, H, C_{in}, C_{out}$ are the filter width, filter height, the number of input channels, and the number of output channels, respectively. The typical way of applying an orthogonality is as follows:
\vspace{-5pt}
\begin{equation}
    \frac{\lambda}{|\mathcal{W}|} \sum_{\mathbf{W} \in \mathcal{W}} \| \mathbf{W}^\top \mathbf{W} - \mathbf{I}_n\|^2_F,
\label{equation3}
\end{equation}
\vspace{-5pt}

\noindent where $\lambda$ is the regularization coefficient, $|\mathcal{W}|$ is the cardinality of $\mathcal{W}$, and $\| \cdot \|_F$ is the Frobenius norm of the matrix.

Recent works\,\cite{wang2020orthogonal,qi2020deep} propose a direct constraint that preserves the norm between the $\mathbf{x}$ and $\mathbf{y}$. Qi et al.\,\cite{qi2020deep} show that the reshape-based method makes a network \textit{isometry} partially because reshaping convolutional filter to 2D is an ad-hoc solution to enable norm preservation property except for when the kernel size and stride of the convolutional filter is equal to 1. In TRADE, we have verified the benefits of the OR for the generalization of architecture candidates in a query phase, as we will see in the experiments.

\section{Method}
\begin{table}[t]
\begin{center}
\footnotesize\addtolength{\tabcolsep}{-3.5pt}
\caption{Top1 accuracy on ImageNet-1k. Each performance is from the pre-trained checkpoint in the PyTorch timm GitHub repository\,\cite{rw2019timm}.\label{table1}} 
\begin{tabular}{c|c|c|c|c}
\toprule
Tag  & Model name            & Input size & Top1 accuracy & NP \\\midrule
T(A) & EfficientnetV2-M      & 224        & 80.2          & 43M        \\\midrule
T(B) & RegNetY-8.0GF\         & 224        & 81.0          & 24M        \\\midrule
T(C) & ViT-Large-16         & 224        & 85.8          & 164M       \\\midrule
T(D) & VIT-Large-32-R50          & 224        & 84.4          & 86M     
\\\bottomrule
\end{tabular}%
\end{center}

\end{table}

\begin{table}[]
\centering\small
\caption{Top1 accuracy on ImageNet-1k for students with different teachers. A student1 is EfficientNetV2-Small and a student2 is RegNetX-3.2GF. The label is annotated in \autoref{table1}.\label{tab:pre-stu-acc}}
\begin{tabular}{c|cc|c}\toprule
Teachers & Student1 & Student2 & Comparison                     \\\midrule\midrule
T(A)     &  \textbf{80.1}  &  78.7  & \cellcolor{LightCyan} \textbf{Student1} \textgreater Student2 \\
T(B)     &  79.8  &  \textbf{80.9} & \cellcolor{LightCyan} Student1 \textless \textbf{Student2}   \\\bottomrule
\end{tabular}
\vspace{-10pt}

\end{table}

\vspace{-5pt}
\begin{figure}[t]
    \centering
    \includegraphics[width=1.0\linewidth]{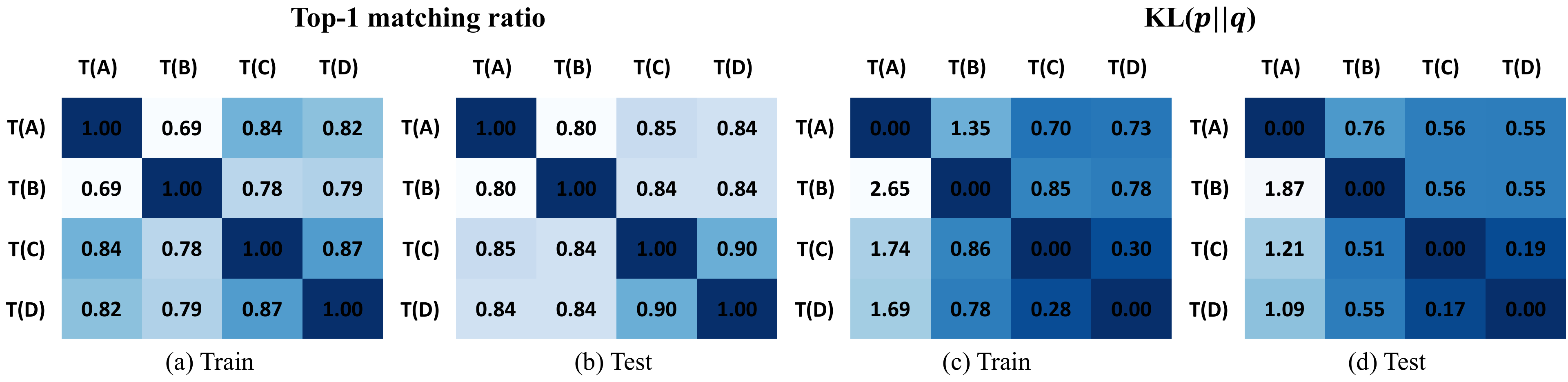}
    \vspace{-15pt}
    \caption{Confusion matrix for models’ outputs, $p$ or $q$ denotes softmax probability output from different architectures. (a), (b): Top-1 matching ratio; (c), (d): KL(p$\|$q) for training and test set. The label is annotated in \autoref{table1}.}
    \label{fig:conf_kd}
    \vspace{-10pt}
\end{figure}

\subsection{Review of Architecture-aware KD\,(AKD)}

AKD\,\cite{liu2020search} searches for student architectures using RL based NAS process with a KD-based reward function with a certain threshold for latency. 
They argue that distilling the same teacher model to different students leads to different performance results, and no pre-defined student architecture produces the best results across all teacher networks. They also have claimed that the optimal student models in AKD can be deemed as the most structure-similar to the teacher model\,(\autoref{tab:pre-stu-acc}). As an extension of previous works, we investigate the teacher-student relationship using recent state-of-the-art architectures\,(\autoref{table1}) consisting of MBConv\,\cite{sandler2018mobilenetv2}, Fused-MBConv\,\cite{tan2021efficientnetv2}, and Transformer block that have not been studied yet. Further from the previous experiment, we obtain a confusion matrix for the training dataset as well as the test dataset because KD is performed by logit matching of training data.

\autoref{fig:conf_kd} shows the confusion matrices of the following evaluation metrics:
\begin{itemize}
    \item \textbf{Top-1 matching ratio $(i,j)$}: accuracy of how well the predicted label matches between $i$-th row model and $j$-th column model\,(\autoref{fig:conf_kd} (a), (b)).
    \item \textbf{Kullback-Leibler\,(KL) divergence $(i,j)$}: the relative entropy of the $i$-th row model's output distribution to that of $j$-th column model\,(\autoref{fig:conf_kd} (c), (d)).
\end{itemize}

\noindent According to the block style, the learning distribution seems to differ significantly, while the Transformer based networks seem to be close to each other. Because these recent architectures have not yet been explored in this direction of research, we aim to redesign search spaces so that all types of blocks are considered. Throughout this work, we expect to know more about architecture-aware KD for any given teacher and to uncover teacher-student relationships.

\section{Understanding Search Efficiency}
\vspace{-5pt}
\paragraph{Fused MB Conv \& MB Conv \& Transformer.}
Recent works try to combine different operators to form a new network\,\cite{xiao2021early,dai2021coatnet,tan2021efficientnetv2}.  
In order to balance (hard) inductive biases and the Transformer blocks' capacity for representation learning, Xiao et al.\,\cite{xiao2021early} hypothesize that restricting convolutions in ViT to early visual processing may be an important design decision. Tan et al.\,\cite{tan2021efficientnetv2} provide the effectiveness of using depthwise convolution; the standard convolutional operator is rather apposite for mobile or server accelerators\,\cite{gupta2020accelerator,li2021searching}. Dai et al.\,\cite{dai2021coatnet} present a key insight regarding hybridizing convolution and attention; simply stacking convolutional and attention layers could be surprisingly effective to achieve better generalization and capacity. Inspired by such findings, we prioritize the combination of the convolutional operators and the Transformer for constructing the architecture while there remains no precedence between Fused-MBConv\,\cite{gupta2019efficientnet} and MBConv\,\cite{sandler2018mobilenetv2}.

\begin{center}\small
    \colorbox{LightCyan}{\textbf{Stem Conv $\rightarrow$ Fused-MBConv \& MBConv $\rightarrow$ Transformer}}
\end{center}

\paragraph{TurBO.}
TurBO\,\cite{eriksson2019scalable} is used to search across the search space for the best architecture. TuRBO utilizes different subregions called trust regions to perform multiple BO. After that, it finds the next promising candidate through implicit Thompson sampling. In our TRADE framework, TurBO proceeds along with the following steps:
\vspace{-5pt}
\begin{enumerate}
    \item Initial architectures are sampled from the Latin hypercube sampling\,\cite{mckay2000comparison}.
    \item Gaussian process is trained with the previous observations. Then, it takes samples of the Sobol sequence\,\cite{sobol1967distribution} from the bounded region and identifies the most promising candidates by computing their likelihood.
    \item It recognizes success and doubles the length of the related trust region if the candidate turns out to be the new best point. In contrast, it sees all other situations as failures and halves them all. The trust region restarts from scratch if the rescaled length reaches a certain threshold.
\end{enumerate}

\vspace{-15pt}
\paragraph{Hyperparameters in TurBO.}
We investigate the optimal hyperparameter settings of TRADE. Referred to Eriksson et al.\,\cite{eriksson2019scalable}, we empirically observe that a larger batch size leads the surrogate model to find better optima. For a quick restarter of TRADE, we close the gap between the minimum threshold and maximum threshold of its hyperrectangle and settle down the tolerance at 2, which controls the trade-off between exploitation and exploration. Implementation details are in the Appendix.

\paragraph{Search Space.}
Similar to \cite{tan2019mnasnet,liu2020search,anonymous2022uninet}, our search space consists of eight pre-defined blocks, with each block containing a list of identical layers. \autoref{tab:block-type} summarizes the usage of various block types in recent architecture methods. Further than the previous methods, we jointly search the optimal combination of convolution blocks and Transformer for new architecture-aware KD by looking at all-operators. Referring to Xiao et al.\,\cite{xiao2021early}, the convolutional blocks are placed in the early stage, and the Transformer block is designed to connect the convolutional block and the head. If Transformer is not used, it is designed to connect the Conv-backbone and head\,(\autoref{fig:overview}). We search for the number of layers, the convolutional block type, the usage of the Transformer blocks, and skip operation type, Conv kernel size, squeeze-and-excite ratio, and input/output filter size for each block independently.

\section{Results}
\begin{figure}[t]
\centering
    \includegraphics[width=1.0\linewidth]{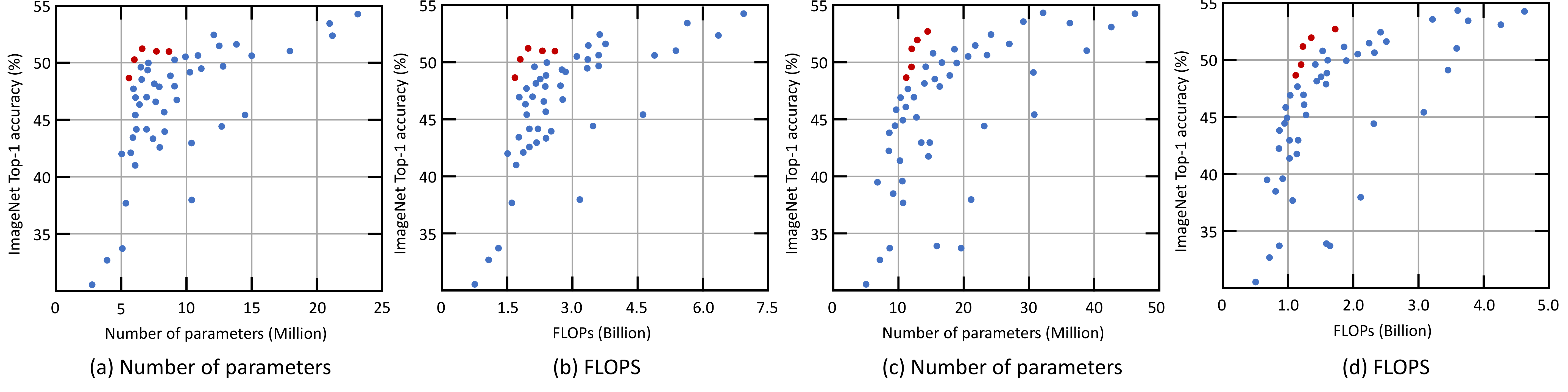}
    \caption{Attributes of searching process. The architecture evolves during searching and each dot represents an architecture. (a), (b): RegNetY-8.0GF\,\cite{radosavovic2020designing} is used as the teacher model; (c), (d): EfficientNetV2-M\,\cite{tan2021efficientnetv2} is used as the teacher model.} \label{fig:pareto-optimal}
\end{figure}

\subsection{Experimental Settings}
Our TRADE process mostly follows the same settings in MNAS\,\cite{tan2019mnasnet} and AKD\,\cite{liu2020search}. We reserve 50K images from the training set as mini-val, and use the rest as mini-train. We treat each sampled model as a student, and distill the teacher’s knowledge by training on the mini-train for 9 epochs, including the first two epochs with warm-up learning rate, and then evaluating on the mini-val to obtain its accuracy. After the aforementioned proxy task, the sampled model will be evaluated on the NVIDIA 2080-Ti GPU to measure its latency. We do not use a shared weight among different architecture candidates.

In each searching experiment, the Gaussian process which is the surrogate model in TurBO samples 50 models. Then we pick the top models that meet the accuracy bracket in Eq.~(\ref{eq:reward}), and train them for further 450 epochs by either distilling the same teacher model or using ground truth labels while applying the OR with the same strategy. Following standard practices in KD, temperature $\tau$ is set to 1, and the balancing hyperparameter $\alpha$ in Eq.~(\ref{equation1}) is set to 0.7. We launch all the searching experiments with 8 NVIDIA 3090-Ti GPUs. Other than those mentioned, we use the same training settings for EfficientNet-EdgeTPU-Small\,\cite{gupta2019efficientnet} in PyTorch timm GitHub\,\cite{rw2019timm}. For better readability, we denote the family of architecture searched by TRADE as TRADE-xxx, whose xxx is the teacher family. For instance, if the ViT-Large-16 is used for the teacher model, then we call the searched architecture the TRADE-ViT. \autoref{fig:pareto-optimal} shows that TRADE finds Pareto-optimal front\,(red dot points) on a population of 50 architectures until convergence. Where applicable, any teacher and hardware measurements can be used in the KD-guided score of the local Bayesian search phase. 

\subsection{Main Results}
In this subsection, we provide the superiority of TRADE and uncover the structural knowledge in KD by answering the following questions referred to by Liu et al.\,\cite{liu2020search}:\vspace{-5pt}

\begin{itemize}
    \item \textbf{Q1)} If TRADE is performed on two different teacher architectures, will they converge to a different area in search phase?
    \item \textbf{Q2)} If TRADE is performed on the same teacher, will they converge to a similar area?

\end{itemize}

\paragraph{A1) Different teachers with the same TRADE.}

We observe that different teachers with different block types lead the student models being comprised of different blocks. \autoref{tab:block_sequence} shows the block sequences of the searched models from TRADE with different teacher models. Specifically, TRADE-ViT only has the Transformer blocks at the end while the other models do not.

\begin{table}[t]
\centering\small
\caption{Block sequences searched from TRADE; FMB: Fused-MBConv, MB: MBConv, T: Transformer.\label{tab:block_sequence}}
\begin{tabular}{c|c}\toprule
Architecture & Block sequence \\\midrule\midrule
TRADE-Reg    & FMB - MB - MB - MB   \\
TRADE-ViT    & FMB - FMB - FMB - MB - MB - T        \\
TRADE-EffNet & MB - FMB - MB - FMB - MB \\\bottomrule
\end{tabular}
\vspace{-10pt}
\end{table}

\vspace{-10pt}
\paragraph{A2) Same teachers with different TRADE.}
We examine whether the searched candidate is produced by the random factor laid in the local Bayesian search. To answer this question, two TRADE algorithms are launched under the same teacher model\,(ViT-large-16 model) and different random seeds. These changes can yield randomness in initial points and local search areas. They converge to a similar area and have a similar design strategy. The different thing is the number of channels and the number of transformer blocks while keeping the block sequence and similar specifications about NP, FLOPS, latency, and top1 accuracy.
\begin{itemize}
    \item Seed 0 : FMB - FMB - FMB - MB - MB - T (x 10)
    \item Seed 1 : FMB - FMB - FMB - MB - MB - T (x 7)
\end{itemize}

\begin{figure}[t]
    \centering
    \includegraphics[width=0.45\linewidth]{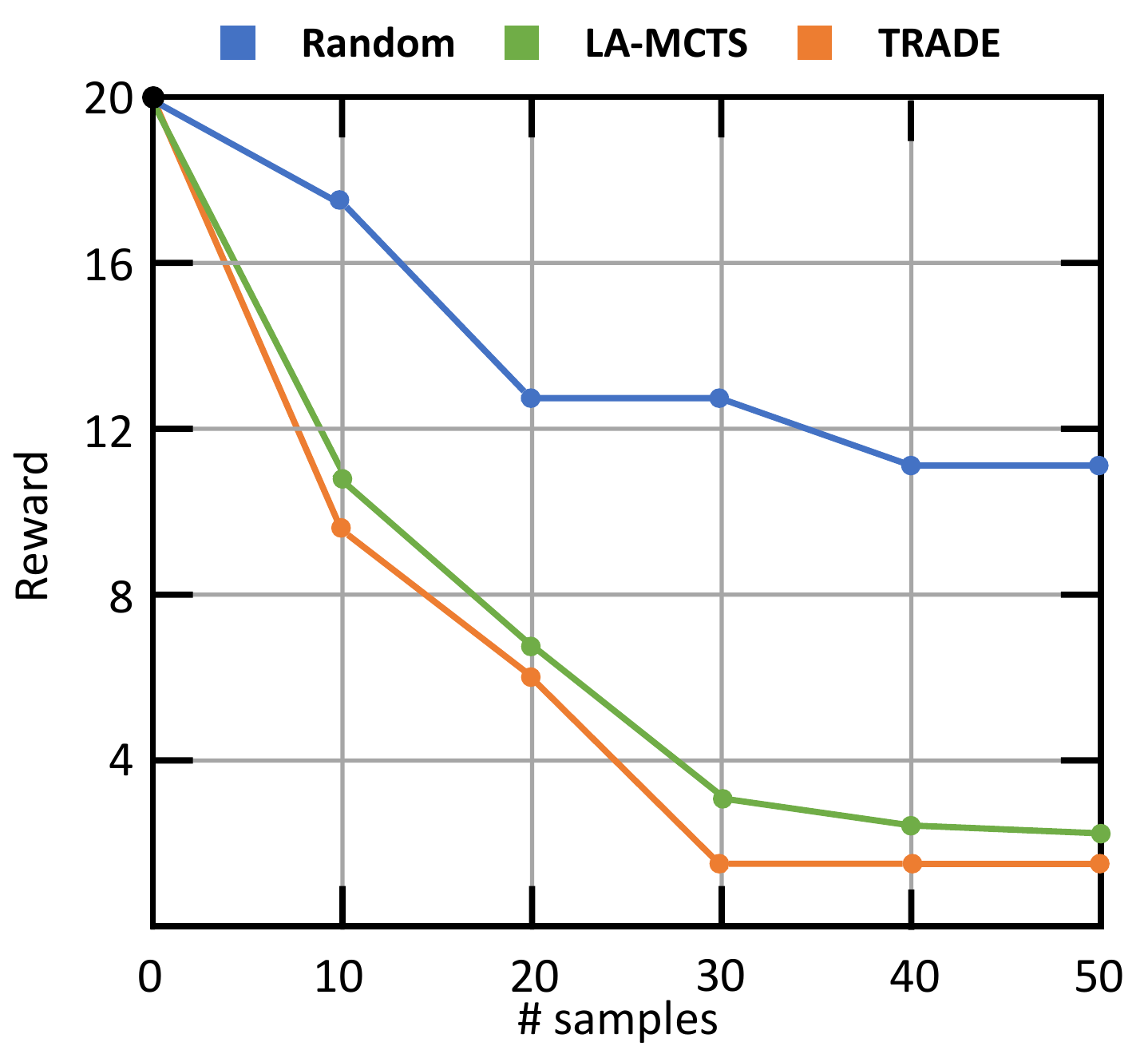}
    \vspace{-5pt}
    \caption{Performance of TRADE against random and LA-MCTS\,\cite{wang2020learning} methods on NAS proxy tasks when ViT-Large-16 model is used for the teacher model.} \vspace{-5pt}
    \label{fig:search_efficacy}
\end{figure}

\vspace{-10pt}

\begin{table}[t]
\centering\small\addtolength{\tabcolsep}{-3pt}
\centering\caption{Performance results on transfer learning datasets.\label{tab:trans}}
\begin{tabular}{cccc} \toprule
Architecture & ImageNet-1k & CIFAR-10 & CIFAR-100 \\\midrule\midrule
TRADE-Reg    &  81.5       &  99.0    &  89.9     \\\midrule
TRADE-ViT    &  81.5       &  99.1    &  88.7     \\\midrule
TRADE-EffNet &  80.3       &  98.8    &  88.8     \\\bottomrule
\end{tabular}
\vspace{-10pt}
\end{table}

\paragraph{Transferability.} For this experiment, we use highlighted checkpoints in \autoref{tab:tea_stu_acc}. Our finetuning settings are mostly the same as ImageNet training with a few modifications similar to \cite{dosovitskiy2020image,touvron2021training,tan2021efficientnetv2}: We use the batch size 256, smaller initial learning rate of 0.001 with cosine decay. For all datasets, we train each model for fixed 20,000 steps. Since each model is finetuned with very few steps, we disable weight decay and use a simple cutout data augmentation. \autoref{tab:trans} shows the transfer learning performance.